\documentclass[11pt]{article}
\usepackage{coling2020}
\usepackage{times}
\usepackage{url}
\usepackage{latexsym}
\usepackage{hyperref}

\usepackage{graphicx}
\usepackage{comment}
\usepackage{array}

\usepackage{times}
\usepackage{latexsym}
\usepackage{comment}
\usepackage{amsmath}
\mathchardef\mhyphen="2D
\usepackage{graphicx}
\graphicspath{ {./} }
\usepackage{float}

\usepackage{multirow}

\usepackage{amsmath}

\usepackage{url}

\colingfinalcopy 

\title{What Does This Acronym Mean? Introducing a New Dataset for Acronym Identification and Disambiguation}

\author{Amir Pouran Ben Veyseh\textsuperscript{\rm 1}, Franck Dernoncourt\textsuperscript{\rm 2}, Quan Hung Tran\textsuperscript{\rm 2}, \\and {\bf Thien Huu Nguyen}\textsuperscript{\rm 1} \\
\textsuperscript{\rm 1} Department of Computer and Information Science, University of Oregon, USA\\
\textsuperscript{\rm 2} Adobe Research, San Jose, CA, USA\\
  \texttt{\{apouranb,thien\}@cs.uoregon.edu}, \\{\tt \{franck.dernoncourt,qtran\}@adobe.com}
}

\date{}

\begin{document}
\maketitle
\begin{abstract}
Acronyms are the short forms of phrases that facilitate conveying lengthy sentences in documents and serve as one of the mainstays of writing. Due to their importance, identifying acronyms and corresponding phrases (i.e., acronym identification (\textbf{AI})) and finding the correct meaning of each acronym (i.e., acronym disambiguation (\textbf{AD})) are crucial for text understanding.
Despite the recent progress on this task, there are some limitations in the existing datasets which hinder further improvement. More specifically, limited size of manually annotated AI datasets or noises in the automatically created acronym identification datasets obstruct designing advanced high-performing acronym identification models. Moreover, the existing datasets are mostly limited to the medical domain and ignore other domains. In order to address these two limitations, we first create a manually annotated large AI dataset for scientific domain. This dataset contains 17,506 sentences which is substantially larger than previous scientific AI datasets. Next, we prepare an AD dataset for scientific domain with 62,441 samples which is significantly larger than previous scientific AD dataset. Our experiments show that the existing state-of-the-art models fall far behind human-level performance on both datasets proposed by this work. In addition, we propose a new deep learning model which utilizes the syntactical structure of the sentence to expand an ambiguous acronym in a sentence. The proposed model outperforms the state-of-the-art models on the new AD dataset, providing a strong baseline for future research on this dataset \footnote{Dataset for AI is available at \href{https://github.com/amirveyseh/AAAI-21-SDU-shared-task-1-AI}{https://github.com/amirveyseh/AAAI-21-SDU-shared-task-1-AI} and dataset for AD is available at \href{https://github.com/amirveyseh/AAAI-21-SDU-shared-task-2-AD}{https://github.com/amirveyseh/AAAI-21-SDU-shared-task-2-AD}}.

\end{abstract}
\section{Introduction}
\label{sec:intro}
Acronyms are shortened forms of a longer phrase. As a running example, in the sentence ``\textit{The main key performance indicator, herein referred to as KPI, is the E2E throughput}" there are two acronyms \textit{KPI} and \textit{E2E}. Also, the acronym \textit{KPI} refers to the phrase \textit{key performance indicator} (a.k.a. the long form of the acronym KPI). In written language, acronyms are prevalent in technical documents that helps to avoid the repetition of long and cumbersome terms, thus saving text space. For instance, about 15\% of PubMed queries include abbreviations,
and about 14.8\% of all tokens in a clinical note
dataset are abbreviations \cite{islamaj2009understanding,xu2007study,jin2019deep}. 

Considering the widespread use of acronyms in texts, a text processing application, such as question answering or document retrieval, should be able to correctly process the acronyms in the text and find their meanings. To this end, two sub-tasks should be solved: 
\textbf{Acronym Identification (AI)}: to find the acronyms and the phrases that have been abbreviated by the acronyms in the document. In the running example, the acronyms \textit{KPI} and \textit{E2E} and the phrase \textit{key performance indicator} should be extracted. \textbf{Acronym Disambiguation (AD)}: to find the right meaning for a given acronym in text. In the running example, the systems should be able to find the right meanings of the two acronyms \textit{KPI} and \textit{E2E}. Note that while the meaning of \textit{KPI} is found in the sentence, the meaning of \textit{E2E} should be inferred from its occurrences either in the previous text of the same document or external resources (e.g., dictionaries). Acronym Identification and disambiguation can be used in many downstream applications including slot-filling \cite{veyseh2019improving}, definition extraction \cite{kang2020document,veyseh2019improving}, and  question answering \cite{ackermann2020resolution,veyseh2016cross}

Over the past two decades, several approaches and resources have been proposed to solve the two sub-tasks for acronyms. These approaches extend from rule-based methods for AI and feature-based models for AD (i.e., SVM and Naive Bayes) \cite{schwartz2002simple,nadeau2005supervised,okazaki2006building,yu2007using} to the recent deep learning methods \cite{li2015acronym,charbonnier2018using,wu2015clinical,ciosici2019unsupervised,jin2019deep,li2019neural}. 
While the prior work has made substantial progress on this task by providing new approaches and datasets, there are some limitations in the existing datasets which hinder further improvement. First, most of the existing datasets for AI are either limited in their sizes or created using simple rule-based methods (i.e., not humnan-annotated). For instance, Ciosici \shortcite{ciosici2019unsupervised} exploits the rules proposed by Schwartz \shortcite{schwartz2002simple} to generate a corpus for acronym disambiguation from Wikipedia. This is unfortunate as rules are in general not able to capture all the diverse forms to express acronyms and their long forms in text \cite{harris2019my}. This limitation not only restricts the coverage of the previous methods and datasets for AI, but also imposes some bias to evaluation of AI using these datasets. Second, most of the existing datasets are in the medical domain, ignoring the challenges in other scientific domains. While there are a few datasets for general domain (e.g., Wikipedia), online forums, news and scientific documents \cite{thakker2017acronym,li2018guess,charbonnier2018using,liu2011learning,harris2019my}, they still suffer from either noisy examples inevitable in the heuristically generated datasets \cite{charbonnier2018using,ciosici2019unsupervised,thakker2017acronym} or their small size, which makes them inappropriate for training advanced methods (e.g., deep neural networks) \cite{prokofyev2013ontology,harris2019my,nautial2014finding}.

In order to address the limitations of the existing resources and push forward the research on acronym identification and disambiguation, especially in scientific domain, this paper introduces two new datasets for AI and AD. Notably, our datasets are annotated by human to achieve high quality and have substantially larger numbers of examples than the existing AI datasets in the non-medical domain (see Appendix \ref{dataset-comparison}). Moreover, based on acronyms and long forms obtained by AI annotations, we introduce a new AD dataset that is also substantially larger than the existing AD datasets for the scientific domain. For more details about the datasets, see Appendices \ref{dataset-statistics} and \ref{dataset-comparison}. Finally, we conduct extensive experiments to evaluate the state-of-the-art methods for AI and AD on the proposed datasets. The experiments show that the existing models fail to match the human-level performance on the proposed datasets.



Motivated by the unsatisfactory performance of the existing AD models, we introduce a new deep learning model for acronym disambiguation. In particular, one of the limitations of the prior AD models is that they fail to efficiently encode the context of the ambiguous acronyms in the given sentences. On the one hand, the traditional feature-based models for AD are limited in their ability to effective representations for the contexts due to the use of hand-designed features. On the other hand, the current deep learning models for AD, based on either language models or sentence encoders (e.g., LSTMs) to capture the contextual information, cannot effectively encode the long dependencies between words in the sentences. These long dependencies are important for acronym disambiguation as the meanings of the acronyms might depend on some words that are sequentially far from the acronyms of interest. To address the issue of long dependencies in sentences, prior work on other natural language processing (NLP) applications has shown that dependency trees can be employed to capture dependencies between words that are sequentially far from each other \cite{veyseh2019improving}. As none of the recent AD models employs the dependency trees to encode the sentence context for acronyms, we propose a novel deep learning model for AD that exploits the syntactic dependency structures of the sentences using graph convolutional neural networks (GCN) \cite{kipf2016semi}. In this work, the structure information induced by the GCN model would be employed as the structural context for acronym disambiguation. Our experiments show that the proposed model outperforms the existing state-of-the-art AD models, providing a strong baseline for the future work on the proposed dataset for AD. 

In summary, in this paper, we make the following contributions:

\begin{itemize}
    \item We release the first publicly available and the largest manually annotated acronym identification dataset in scientific domain. Moreover, we also release the largest acronym disambiguation dataset in scientific domain which is created from humanly curated dictionary of ambiguous acronyms
    \item We conduct extensive experiments on the proposed datasets and compare the performance of the state-of-the-art acronym identification and disambiguation systems.
    \item We propose a new graph-based acronym disambiguation method that employs the syntactic structure of the sentence to predict the expanded form of the ambiguous acronym. This baseline outperforms existing state-of-the-art models for acronym disambiguation. 
\end{itemize}


\section{Data Collection and Annotation}
In order to prepare a corpus for acronym annotation, we collect a corpus of 6,786 English papers from arXiv. These papers consist of 2,031,592 sentences
that would be used for data annotation for AI and AD in this work.


\subsection{AI Dataset}
AI aims to identify acronyms (i.e., short forms) and phrases that have been abbreviated by the acronyms in text (i.e., long forms). To create a dataset for this task, we manually annotate the sentences from the collected papers. In order to improve the efficiency of the data annotation, we seek to annotate only a set of sentences $S=\{s_1,s_2,...,s_n\}$ that have a high chance to host short forms or long forms for acronyms. As such, each sentence $s_i$ for annotation needs to contain at least one word $w_a$ in which more than half of the characters in $w_a$ are capital letters (i.e., acronym candidates). Afterward, for each sentence $s_i$ in $S$, we search for a sub-sequence of words $W_t=\{w_t,w_{t+1},...,w_{t+k}\}$ in $s_i$ in which the concatenation of the first one, two or three characters of the words $w_j \in W_t$ (in the order of the words in the sub-sequence $W_t$) could form an acronym candidate $w_a$ in $s_i$\footnote{Note that we also allow some words $w_j \in W$ to be omitted when we make the concatenation of the characters. Also, note that $w_a$ could be any acronym candidate in $s_i$}.
We call $W_t$ as long form candidate. If we cannot find any long form candidate $W_t$ in sentence $s_i$, we remove sentence $s_i$ from $S$.
Using this process, we end up with 17,506 sentences in $S$ to be annotated manually by the annotators from Amazon Mechanical Turk (MTurk). In particular, we create a HIT for each sentence and ask the workers to annotate the short forms and the long forms in the sentence. Also, we ask them to map each long form to its corresponding short form. In order to work on these HITs, workers should have an approval rate of more than 99\% and pass a qualification test. Moreover, we remove the annotations from workers who submit the HIT sooner than a preset time or have more than 10\% of their annotations different from with the others'. We hire three workers per HIT and pay each of them \$0.05. In case of disagreements, if two out of three workers agree on an annotation, we use majority voting to decide the correct annotation. Otherwise, a fourth annotator is hired to resolve the conflict. The inter-annotator agreement (IAA) using  Krippendorff’s alpha \cite{krippendorff2011computing} with the MASI distance metric \cite{passonneau2006measuring} for short-forms (i.e., acronyms) is 0.80 and for long-forms (i.e., phrases) is 0.86. Out of the 17,506 annotated sentences, 1\% and 24\% of the sentences do not contain any short form or long form, respectively. We call this dataset \textbf{SciAI} (scientific acronym identification). In this work, the AI task is formulated as sequence labeling where we provide the boundaries for the acronyms and long forms in the sentence using BIO format (i.e., label set includes \textit{B-acronym}, \textit{I-acronym}, \textit{B-long}, \textit{I-long} and \textit{O}).


\subsection{AD dataset}
\label{sec:ADDATA}
Using the AI dataset annotated in the previous section, we first create a dictionary of acronym meanings (i.e., a mapping from short forms to all of the possible long forms). First, we create a meaning candidate set\footnote{In this paper, we use \textit{meaning} and \textit{long form} interchangeably} for each acronym based on the annotations in AI dataset. As some of the long forms in the meaning set might be different variants of a single meaning, we need to normalize the long forms of each acronym to find the canonical set of long forms of the acronym. To achieve this goal, we use the normalization approach proposed by the prior work \cite{ciosici2019unsupervised}. The main idea is to   compute the  Levenshtein string edit distance between any pair of long forms, and if it is smaller than a threshold, we replace the less common long form with the more frequent one. After creating the canonical set of long forms for the acronyms, we remove unambiguous acronyms from the mapping/dictionary (i.e., acronyms with only one possible long form), leading to a dictionary $\mathcal{D}$ of ambiguous acronyms. Furthermore, in order to improve the quality of the mappings in $\mathcal{D}$, a graduate student curates the dictionary $\mathcal{D}$ to verify the canonical forms and resolve any potential error. Afterwards, we employ $\mathcal{D}$ to create an AD dataset. To this end, we assume that if an acronym is defined in a paper, its meaning will not change throughout that paper, i.e., the one-sense-per-discourse assumption. Using this criterion, for each acronym $a$ in the dictionary $\mathcal{D}$, we look up all sentences $s_i$ in $S$, i.e., the set of sentences annotated for AI, that contains $a$ and its corresponding long form $l$ (indicated by the annotation for SciAI). Note that here $l$, or its more frequent variant, is already in the canonical long form set of $a$ in $\mathcal{D}$. Then, we assign the long form $l$ to all the occurrences of $a$ in the document that contains $s_i$. This approach leads to a dataset of 62,441 samples where each sample involves a sentence, an ambiguous acronym, and its correct meaning (i.e., one of the meanings of the acronym recorded by the dictionary $\mathcal{D}$). We call this dataset \textbf{SciAD} (scientific acronym disambiguation). Consequently, AD is formulated as a classification problem where given a sentence and an acronym, we need to predict the long form of the acronym in a given candidate set.

\section{Model}
In this section we describe the details of the proposed model for AD. As mentioned in section \ref{sec:ADDATA}, we formulate this task as sequence classification problem. Formally, given an input sentence $W=w_1,w_2,...,w_n$ and the position of the acronym, i.e., $p$, the goal is to disambiguate the acronym $w_p$, that is, predicting the long form $l$ from all the possible long forms of $w_p$ specified in the dictionary $\mathcal{D}$. The proposed model consists of three major components: (1) \textbf{Sentence Encoder}: This component encodes the input sentence into a sequence of representation vectors for the words, (2) \textbf{Context Encoder}: In order to augment the representation of the acronym $w_p$, the context encoder component is designed to leverage the syntactical structure of the sentence to generate the context representation vector for $w_p$, and (3) \textbf{Prediction}: This component consumes the representations obtained from previous components to predict the long form $l'$ for the acronym $w_p$ in the sentence. In this section, we describes these components in details.

\subsection{Sentence Encoder}

We represent the word $w_i$ of $W$ using its corresponding pre-trained word embedding. Furthermore, the word embedding $w_i$ is concatenated with the POS tag embedding of $w_i$ to obtain the representation vector $e_i$ for $w_i$. In order to create more abstract representations of the words and to encode the sequential order of the words in the sentence, we exploit the popular recurrent architecture encoder (i.e., Bidirectional Long Short-Term Memory (BiLSTM)) to consume the word representations $E=e_1,e_2,\ldots,e_n$ to generate abstract representations $H=h_1,h_2,\ldots,h_n$. More specifically, the abstract representation $h_i$ is obtained by concatenating the corresponding hidden states of the forward and backward LSTMs: $h_i=[\overrightarrow{h_i}:\overleftarrow{h_i}]$. The abstract representation $H$ will be consumed by the next components in our model.

\subsection{Context Encoder}

While the representation generated by recurrent encoders such as BiLSTM could potentially capture the context of the entire sentence to represent each word $w_i$, in practice, these models fail to encode long dependencies due to vanishing gradient issue. This is problematic as the actual meaning of the words, especially the acronym $w_p$, might depends on some words appeared in far distances from the word itself. In order to address this limitation, prior work in other NLP tasks has shown that the syntactical structure (e.g., dependency trees) could substantially improve the range of contextual information encoded in word representations. Unfortunately, none of the existing deep learning models for AD exploits the dependency trees of the sentences to enrich the contextual information encoded in the word representations. Thus, in this work, we propose to leverage the syntactical structure available in the dependency tree to augment the representations of the words with syntactical contextual information. More specifically, the dependency tree of $W$ is modeled as an undirected graph with adjacency matrix $A$ where $A_{i,j}$ is 1 if the word $w_i$ is the head or one of the children of the word $w_j$ in the dependency tree. Note that we add self-loops to the dependency tree by adding the identity matrix $I$ to $A$: $\hat{A}=A+I$. Afterwards, in order to encode the structure $A$ into the node (i.e., word) representations, we employ graph convolutional neural networks (GCN) \cite{kipf2016semi,Pouran:19}. Concretely, we first initialize the node representations with the vectors $H$ generated by BiLSTM and then at the $m$-th layer of the GCN, we update the representation of $i$-th node by: $    h^m_i = \sigma (W_m\cdot\frac{1}{\text{deg}(i)}\Sigma_{j\in\mathcal{N}(i)}h^{m-1}_j)$, where $W_m$ is the weight matrix at layer $m$, $\sigma$ is a non-linear activation function and $\mathcal{N}(i)$ and $\text{deg}(i)$ are the set of neighbors and the degree of the $i$-th node in the adjacency matrix $\hat{A}$, respectively. The word representations at the final layer of the GCN, i.e., $H^s=h^s_1,h^s_2,\ldots,h^s_n$, are then employed as the structure-aware representations to be consumed by the next component.


\subsection{Prediction}

The prediction component is the last layer in our model which employs the representation obtained from the BiLSTM, i.e., $H$, and the GCN, i.e., $H^s$, to predict the true long form of the acronym $w_p$ in sentence $W$. To this end, the representation of the acronym $w_p$ from BiLSTM, i.e., $h_i$, and GCN, i.e., $h^c_i$, are concatenated to capture both the sequential and structural context of the word $w_p$. In addition, we also concatenate the representations of the entire sentence obtained from BiLSTM and GCN to enrich the acronym representation. To create the sentence representations, we employ max pooling over all words: $H_{sent}=MAX\_POOL(h_1,h_2,...,h_n)$ and $H^s_{sent}=MAX\_POOL(h^s_1,h^s_2,...,h^s_n)$. Thus, the final representation vector for prediction will be: $V=[h_p:h^s_p:H_{sent}:H^s_{sent}]$, where ``$:$" indicates concatenation. Finally, a two-layer feed forward classifier is employed to predict long form $l'$. Note that the number of neurons in the last layer of the feed forward classifier is equal to the total number of long forms of all acronyms in $\mathcal{D}$.  We use negative log-likelihood as the loss function to train the model: $L=-P(l|W,p)$. We call this model graph-based acronym disambiguation (GAD).

\section{Experiments}

\begin{table}[]
    \centering
    \begin{tabular}{l|c|c|c}
         & SciAD & UAD & SciUAD \\
         \hline
        Number of acronyms & 732 & 567 & 538 \\ 
        average number of long form per acronym & 3.1 & 2.3 & 1.8 \\
        overlap between sentence and long forms & 0.32 & 0.01 & 0.31 \\
        average sentence length & 30.7 & 18.55 & 50.30 
    \end{tabular}
    \caption{Comparison of different AD datasets. Note that the third row shows the ratio of sentences that have at least one word in common with the long forms of the acronyms appearing in the sentence.
    }
    \label{tab:AD_datasets}
\end{table}

\textbf{Datasets \& Evaluation Metrics}:
We evaluate the performance of the state-of-the-art acronym identification and disambiguation models on SciAI and SciAD, respectively. For AD, in addition to SciAD, we evaluate the performance of the models on the UAD Wikipedia dataset proposed by Ciosici \shortcite{ciosici2019unsupervised}. UAD consists of sentences in general domain (i.e., Wikipedia). As the domain difference between SciAD and UAD (i.e., scientific papers vs Wikipedia articles) could effect on the direct comparison between the performances of the models, we also prepare SciUAD dataset for acronym disambiguation\footnote{Note that we do not use NOA dataset \cite{charbonnier2018using} as it has only 4 samples per each long form on average and it is not suitable for supervised baselines. So, it cannot be comparable with the other datasets in our experiments.}.  SciUAD employs the same corpus as we use to create SciAD but the acronyms, long forms, their mappings, and the ambiguous use of acronyms in corpus are all extracted using the unsupervised method proposed by UAD \cite{ciosici2019unsupervised}. Table \ref{tab:AD_datasets} compares all AD datasets using different criteria. There are several observations from this table. First, SciAD supports more ambiguous acronyms and the level of ambiguity is higher in SciAD as there are more long forms per acronym. This is significant, specially considering the fact that SciAD is prepared from a smaller corpus than UAD. Also, comparing the level of ambiguity in SciAD and SciUAD indicates that our dataset preparation is more effective to capture acronym ambiguity in the given corpus than UAD. Second, comparing the ratio of sentences that have overlap with the long forms emphasizes the difference between scientific domain and general domain. The higher overlap ratio of SciAD and SciUAD compared to UAD could be attributed to two characteristics of the scientific domain: 1) Sentences in scientific papers are normally longer than sentences of general articles and it is corroborated by comparison between average sentence length in the three datasets shown in Table \ref{tab:AD_datasets}. 2) Long forms in scientific domain are more semantically related to each other, therefore they share more vocabulary in the context of their acronyms. This shows that AD could be more challenging on scientific domain than general domain. For more statistics, see Appendix \ref{dataset-statistics}.

 To create training and testing splits of the datasets, we randomly divide SciAI, SciAD and UAD into training, development and test data using 80:10:10 ratio. Regarding the evaluation metrics, for AI, a prediction of acronym or long form is counted as true if the boundaries of the prediction matches with the boundary of the ground-truth acronym or long form in the sentence, respectively. We report the macro-averaged precision, recall and F1 score computed for acronym and long form prediction. For AD, similar to prior work \cite{ciosici2019unsupervised}, we report the performance of the models using macro-averaged precision, recall and F1 score computed for each long form. 

\begin{table}[]
    \centering
    \resizebox{.58\textwidth}{!}{
    \begin{tabular}{c|ccc|ccc|c}
        Model & \multicolumn{3}{c}{Acronym} & \multicolumn{3}{c}{Long Form} & \\ \hline
        & P & R & F1 & P & R & F1 & Macro F1 \\ \hline
        NOA & 80.31 & 18.08 & 29.51 & 88.97 & 14.01 & 24.20 & 26.85 \\
        ADE & 79.28 & 86.13 & 82.57 & 98.36 & 57.34 & 72.45 & 79.37 \\
        UAD & 86.11 & 91.48 & 88.72 & 96.51 & 64.38 & 77.24 & 84.09 \\ \hline \hline
        BIOADI & 83.11 & 87.21 & 85.11 & 90.43 & 73.79 & 77.49 & 82.35 \\
        LNCRF & 84.51 & 90.45 & 87.37 & 95.13 & 69.18 & 80.10 & 83.73 \\ \hline \hline
        LSTM-CRF & 88.58 & 86.93 & 87.75 & 85.33 & 85.38 & 85.36 & 86.55 \\ \hline \hline
        Human Performance & 98.51 & 94.33 & 96.37 & 96.89 & 94.79 & 95.82 & 96.09
    \end{tabular}
    }
    \caption{Performance of models in acronym identification (AI)}
    \label{tab:AI}
\end{table}

\begin{table}[]
    \centering
    \resizebox{.68\textwidth}{!}{
    \begin{tabular}{c|ccc|ccc|ccc}
        Model & \multicolumn{3}{c}{SciAD} & \multicolumn{3}{c}{UAD} & \multicolumn{3}{c}{SciUAD} \\ \hline
        & P & R & F1 & P & R & F1 & P & R & F1 \\ \hline
        MF & 89.03 & 42.2 & 57.26 & 76.37 & 46.34 & 57.68 & 91.32 & 45.21 & 60.47 \\
        ADE & 86.74 & 43.25 & 57.72 & 83.56 & 44.01 & 57.65 & 85.90 & 42.57 & 56.92 \\
        \hline \hline
        NOA & 78.14 & 35.06 & 48.40 & 76.93 & 42.79 & 54.99 & 79.23 & 36.76 & 50.21 \\
        UAD & 89.01 & 70.08 & 78.37 & 90.82 & 92.33 & 91.03 & 90.23 & 72.43 & 80.35 \\
        BEM & 86.75 & 35.94 & 50.82 & 75.33 & 44.52 & 55.96 & 85.99 & 37.24 & 51.97 \\
        DECBAE & 88.67 & 74.32 & 80.86 & 95.23 & 93.74 & 94.48 & 90.11 & 75.13 & 81.94 \\ \hline \hline
        GAD & 89.27 & 76.66 & 81.90 & 96.06 & 94.37 & 95.21 & 91.12 & 77.08 & 83.51 \\ \hline \hline
        Human Performance & 97.82 & 94.45 & 96.10 & 98.13 & 96.32 & 97.21 & 98.02 & 95.43 & 96.70
    \end{tabular}
    }
    \caption{Performance of models in acronym disambiguation (AD)}
    \label{tab:AD}
\end{table}

\textbf{Baselines}:
For acronym identification, we compare the performance of the following baselines on SciAI: (1) \textbf{Rule-based methods}: These models employ manually designed rules and regular expressions to extract acronyms and long forms in text; namely we report the performance of NOA \cite{charbonnier2018using}, UAD \cite{ciosici2019unsupervised} \footnote{Note that UAD employs the rules proposed by \cite{schwartz2002simple}} and ADE \cite{li2018guess}, (2) \textbf{Feature-based models}: These models define a set of features for acronym and long form predictions, then they train a classifier (e.g., SVM or CRF) using these features, namely we report the performance of BIOADI \cite{kuo2009bioadi} and LNCRF \cite{liu2017multi}, (3) \textbf{Deep learning models}: Since to the best of our knowledge, none of the existing work leverage pre-trained word embeddings with deep architectures for AI, we implement an LSTM-CRF baseline for this task. For more details on the LSTM-CRF baseline and hyper parameters, see Appendices \ref{sec:lstmcrf} and \ref{sec:parameters}.

For acronym disambiguation, we compare the performance of the proposed model, i.e., GAD, with the following baselines: (1) \textbf{Non-deep learning models}: This category includes two models: (a) ``most frequent'' (MF) which takes the long form with the highest frequency among all possible meanings of an acronym as the expanded form of the acronym, and (b) a feature-based model that employs hand crafted features from the context of the acronyms to train a disambiguation classifier; namely ADE \cite{li2018guess}, (2) \textbf{Deep learning models}: In this category, we report the performance of (1) language-model-based baselines that train the word embeddings using the training corpus, namely NOA \cite{charbonnier2018using} and UAD \cite{ciosici2019unsupervised},  and (2) models employing deep architectures (e.g., LSTM), namely we report performance of DECBAE \cite{jin2019deep} and BEM \cite{blevins2020moving}\footnote{Consider that BEM is originally proposed for word sense disambiguation. As there is no glossary for the long forms, we use the words of the long form as the its glossary}.

Table \ref{tab:AI} and \ref{tab:AD} show the results. There are several important observations from these tables. First, for AI, all rule-based and feature-based methods have higher precision for long form prediction than those for acronym prediction. This is due to the conservative nature of the rules/features exploited for finding long forms. On the other hand, these rules/features fail to capture all patterns of expressing the meanings of the acronym, resulting in poorer recall on long forms compared to acronyms. This is in contrast to the deep learning model LSTM-CRF as it has comparable recall on long forms and acronyms, showing the importance of pre-trained word embeddings and deep architectures for AI. Moreover, the performance of all AI baselines fall far behind human level performance, indicating that more research is required to fill this gap. Second, for Acronym Disambiguation task, except for MF and ADE, all baselines perform substantially better on UAD than SciAD; showing that SciAD is more challenging than UAD. In addition, while the better performance of the baselines on UAD than those on SciUAD corroborates the more challenging nature of AD in the scientific domain, the better performance of the baselines on ScieUAD than those on SciAD shows that the manual annotation for the AD dataset in this paper is more effective than those in UAD. Third, among all the baselines, the proposed GAD achieves the best results on all three datasets, showing the importance of syntactic structure for AD. However, all baselines are still far less effective than human on SciAD, thereby providing many research opportunities on this dataset. 



\section{Analysis}

This sections provides further analysis on the annotation process and the proposed datasets and model. We first discuss the common errors in annotations that result in conflicts. Afterwards, we provide a sample complexity analysis of the AD task. Case study is presented in the end.


\subsection{Annotation Errors}

\begin{table}[]
\begin{center}
\resizebox{.85\textwidth}{!}{
\begin{tabular}{l|c|c}
Error Type & Share & Examples \\ \hline
\multirow{2}{*}{Popular Acronyms} & \multirow{2}{*}{16\%} & ... perspective projection or \textbf{3D} Thin Plate Spline (TPS) ... \\\cline{3-3}
    & & ... the temporal resolution (TR) is 720 \textbf{ms} \\ \hline
\multirow{2}{*}{Meaningful Acronyms} & \multirow{2}{*}{8\%} & ... the \textbf{Cost} library and the \textbf{Sense} simulator ... \\\cline{3-3}
    & & For the \textbf{Home} dataset, we compare simplified ... \\ \hline
\multirow{2}{*}{Embedded Acronyms} & \multirow{2}{*}{18\%} & ... translation on \textbf{Europarl} (EP) and News domains ... \\\cline{3-3}
    & & ... linear \textbf{SVM} (LSVM) and Radial Basis \textbf{SVM} (RSVM) classifiers ... \\ \hline
{Lack of Expertise} & {5\%} & ... current \textbf{malware landscape} and allow re-training the \textbf{ML} systems ... \\ \hline
{Multiple Acronyms} & {24\%} & Since online \textbf{API} access could ... in the \textbf{API} access \\ \hline
\end{tabular}
}
\end{center}
\caption{Errors in Annotations. Words shown in bold are missed from annotation except for \textit{Lack of Expertise} error, where the acronym is mapped to a wrong long form }
\label{tab:errors}
\end{table}

In order to prevent conflicts in annotations, we first conduct a pilot study to analyze the common errors that annotators could make during the annotations, then we update the annotation instructions and interface accordingly to prevent these errors. More specifically, we create HITs for 5\% of the sentences in the prepared corpus and we manually analyze the conflicts. Table~\ref{tab:errors} shows the most frequent errors by the annotators in the annotation process that create the conflicts. Note that these errors represent 75\% of the conflicts and the other 25\% of the conflicts are due to mis-annotations with other errors (e.g., careless annotation). Among all errors, multiple acronyms (i.e., failing to annotate all occurrences of an acronym in a sentence), embedded acronyms (acronyms that are part of a long form),
and popular acronyms (i.e., acronyms such as 3D (3 dimensions) or ms (millisecond)) contribute the most to the conflicts. To reduce these errors and prevent conflicts, we update the instructions in the annotation interface to explicitly address confusions that result in these errors and provide examples to the annotators. In addition to that, we include test cases prone to each of these errors in the qualification test. Based on our evaluations, compared to the pilot study, these modifications result in 43\% reduction in conflict rates during the main annotation process. Details of the annotation instructions and interface are provided in Appendix \ref{sec:interface}. 

\subsection{Sample Complexity Analysis}
\label{sec:sample}
\begin{table}[]
    \centering
    \resizebox{.68\textwidth}{!}{
    \begin{tabular}{c|ccc|ccc|ccc}
        Model & \multicolumn{3}{c}{SciAD} & \multicolumn{3}{c}{UAD} & \multicolumn{3}{c}{UAD$_{small}$} \\ \hline
        & P & R & F1 & P & R & F1 & P & R & F1 \\ \hline
        DECBAE & 88.67 & 74.32 & 80.86 & 95.23 & 93.74 & 94.48 & 89.01 & 93.66 & 91.28 \\
        GAD & 89.27 & 76.66 & 81.90 & 96.06 & 94.37 & 95.21 & 90.78 & 94.59 & 92.64 \\
    \end{tabular}
    }
    \caption{Performance of models on the small version of UAD compared to the original UAD and SciAD daataset in AD task}
    \label{tab:complex}
\end{table}

As it is shown in Table \ref{tab:statistics} in Appendix \ref{dataset-statistics}, UAD dataset is almost one order of magnitude larger than SciAD. This larger size results in more training samples per meaning. More specifically, in SciAD, there are 17 training samples per each long form, while this number is 126 for UAD. So, in this analysis, we tend to answer the question that whether the better results of the baselines on UAD is due to its larger size or its less challenging nature than SciAD. To this end, we prepare a small version of the UAD dataset by keeping only 17 training samples for each long form. Using this criterion, the size of the UAD training set reduces to 22,522. We name this datase as UAD$_{small}$. Note that we do not change the size of development or test set. Afterwards, we retrain the DECBAE and GAD models on the new small UAD training set. The results of evaluation of re-trained models on UAD test set are shown in Table \ref{tab:complex}. For convenience, we also report the results on the original UAD and SciAD datasets. This table shows that training on the UAD$_{small}$ results in performance loss on UAD test set, however, both models still perform considerably better on UAD$_{small}$ than SciAD. It empirically proves that SciAD is more challenging than UAD as the same models trained on datasets with comparable size achieve higher performance on UAD$_{small}$ than SciAD. In addition, this table also shows that in the small version of UAD, i.e., UAD$_{small}$, GAD again outperforms DECBAE, indicating its superiority for AD.

\subsection{Case Study}
\begin{table}[]
\begin{center}
\resizebox{.95\textwidth}{!}{
\begin{tabular}{l|c|c}
Sentence &  Long Form extracted by Rules & Model \\ \hline
It uses \textbf{fast and factual exploration} (\textbf{FA2E}) to perform ... & factual exploration & UAD \\\hline
results from \textbf{overloaded homogeneous preparation} (\textbf{OMP}) for ... & homogeneous preparation & UAD \\\hline
 we first use \textbf{complete polar evaluation} (\textbf{CARE}) as ... & N/A & ADE \\\hline
 which uses \textbf{local rewriting pattern} (\textbf{LWP}) features ... & N/A & ADE \\ \hline
 ... where \textbf{genetic algorithms} come into picture, denoted by \textbf{GA} in ...& N/A & UAD/ADE
\end{tabular}
}
\end{center}
\caption{Failures of rule-based methods for extracting acronyms and their long forms from text}
\label{tab:rule-failurs}
\end{table}

In this section we study the cases in which the acronym identification baselines fail to correctly extract the long form of the identified acronyms. Studying this type of failure is important because it contributes the most to the low recall of the baselines for long form identification. Note that most of the recent work employs rule-based methods for long form identification, so we analyze the failures of two baselines UAD \cite{ciosici2019unsupervised} and ADE \cite{li2018guess}. The results are shown in Table \ref{tab:rule-failurs}. Note that UAD looks for a sequence of letters in the words preceding the acronym which can make the capital letters in the acronym. Thus, it fails in the first two examples. The failures of the ADE are mainly due to the use of characters in the middle or end of the words in the long forms to create the acronyms. In the last example, both methods fail as the long form is far away from the acronym so they do not capture it.

In order to provide more insight into the acronym disambiguation challenges in SciAD, we study the failures and successes of the proposed model GAD. For the success cases, we study sentences in which the other baselines fail to correctly disambiguate the given acronym but GAD can successfully predict the long form. Consider this sentence: ``\textit{Words that are not compatible with our pre-defined rules are excluded from \textbf{SDP}}". In this example, the true long form of ``\textit{SDP}" is ``\textit{shortest dependency path}" and the token ``\textit{Words}" in the beginning of the sentence is a clue to disambiguate this acronym. Due to the long distance between ``\textit{SDP}" and ``\textit{Words}", the baselines that encode the sentence sequentially fail to capture their connection. On the other hand, as GAD is equipped with the dependency tree and it benefits from the short dependency path between ``\textit{SDP}" and ``\textit{Words}". As another example, consider this sentence: ``\textit{\textbf{SDP}, SP, and GT together with the models proposed in section 4  and the simplified version of SLD are the major optimization baselines in our experiments}". In this example, the expanded form of ``\textit{SDP}" is ``\textit{semi-definite programming}". While the other baselines fail to correctly predict the true long form, GAD successfully predicts it. The success of the GAD could be attributed to the short distance between ``\textit{SDP}" and the word ``\textit{optimization}" in the dependency tree (see Figure \ref{fig:dep}). On the other hand, due to the long sequential distance between ``\textit{SDP}" and ``\textit{optimization}" in the sentence, none of the other baselines is able to capture the connection between these two words.

Despite the improvement obtained from utilizing dependency tree for acronym disambiguation, there are still cases that some misleading words in the sentence could obfuscate disambiguation. Note that in these examples both GAD and the baselines fail, indicating the necessity of more advanced models for this task. 
For instance consider this example: ``\textit{$\theta$ represents the parameters that determine the optimal splitting node of \textbf{RF} in our regression model}". In this sentence the true long form of the acronym ``\textit{RF}" is ``\textit{Random Forest}" but the GAD model wrongly predicts the long form ``\textit{Regression Functions}". This failure could be attributed to the existence of the misleading word ``\textit{regression}" in the sentence that highly correlates to the vocabulary of the contexts of the long form ``\textit{Regression Functions}" 
This high correlation downgrades the importance of the related word ``\textit{node}" which is useful to disambiguate this acronym. 

\section{Related Work}

In the last two decades, several work has been proposed for acronym identification and disambiguation. For AI, most of the prior work employs rule-based methods \cite{park2001hybrid,wren2002heuristics,schwartz2002simple,adar2004sarad,nadeau2005supervised,ao2005alice,kirchhoff2016unsupervised} or feature engineering \cite{kuo2009bioadi,liu2017multi}. Majority of the recent work leverages the rules proposed by \cite{schwartz2002simple} to extract long forms from the sentences containing acronyms. Recently, it has been shown that the rule-based methods fall far behind human level performance, especially in scientific domain \cite{harris2019my}. In addition to the rule-based methods, some work utilizes users' search experience to identify the meanings of acronyms \cite{jain2007acronym,nadeau2005supervised,taneva2013mining}. However, this method cannot be applied to non-web-based resources (e..g, scientific papers). For acronym disambiguation, prior work employs either feature-based models \cite{wang2016clinical,li2018guess} or deep learning models. While some of the deep learning models directly employ word embeddings for disambiguation \cite{wu2015clinical,antunes2017biomedical,charbonnier2018using,ciosici2019unsupervised}, some of them employ deep architectures to encode the context of the acronym \cite{jin2019deep,li2019neural}. Moreover, acronym disambiguation has been also modeled as the more general tasks Word Sense Disambiguation (WSD) \cite{henry2017evaluating,tulkens2016using} or Entity Linking (EL) \cite{cheng2013relational,li2015acronym}. While the majority of the prior work studies AD in medical domain \cite{okazaki2006building,vo2016abbreviation,wu2017long}, recently some work proposes acronym disambiguation in general \cite{ciosici2019unsupervised}, enterprise \cite{li2018guess}, or scientific domain \cite{charbonnier2018using}. 


\section{Conclusion}
We introduce new datasets and strong baselines for AI and AD sub-tasks. These datasets are created based on human annotation and shown to be more challenging than their previous counterparts. While the baselines achieve good performance, especially the proposed GAD model which outperforms the prior models for AD, more research is required to reach to human level performance for the new datasets.

\bibliographystyle{coling}
\bibliography{coling2020}

\clearpage

\appendix

\section{LSTM-CRF Baseline}
\label{sec:lstmcrf}
This section describes the details of the LSTM-CRF employed in acronym identification experiments. Formally, given the input sentence $W=w_1,w_2,...,w_n$, each word $w_i$ is represented by $e_i$ which is the concatenation of (1) the corresponding pre-trained word embedding, and (2) the embedding of the part-of-speech (POS) tag of the word $w_i$. Afterwards, the word representations $E=e_1,e_2,...,e_n$ are fed into a bi-directional long short-term memory (BiLSTM) network to obtain more abstract representations $H=h_1,h_2,...,h_n$. Specifically, the representation $h_i$ is obtained by the concatenation of the hidden states of the forward and backward LSTMs at the corresponding time step. Next, the feature vector $h_i$ is transformed into a score vector $v_i$ whose dimensions correspond to the possible word labels/tags (i.e., the five BIO tags) and quantify the possibility for $w_i$ to receive the corresponding labels: $v_i = W_S h_i$, where $W_S$ is the trainable weight matrix and $|v_i| = 5$. In the next step, to capture the sequential dependencies between word labels, we exploit conditional random field (CRF) layer. Concretely, the score for a possible label sequence $\hat{L}=\hat{l}_1, \hat{l}_2, ... , \hat{l}_N$ for $W$ would be:
\begin{equation}
    Score(\hat{l}_1,\hat{l}_2,...,\hat{l}_N|W) = \Sigma_{j=1}^N \big( v_{\hat{l}_j} + t_{\hat{l}_{j-1},\hat{l}_{j}} \big)
\end{equation}

where $t_{\hat{l}_{j-1},\hat{l}_{j}}$ is the trainable transition score from label $\hat{l}_{j-1}$ to label $\hat{l}_{j}$. Note that the probability of the possible labeling $\hat{L}$, i.e., $P(\hat{l}_1,\hat{l}_2,...,\hat{l}_N|W)$, is computed using dynamic programming \cite{Lafferty:01}. We use the negative log-likelihood $\mathcal{L}$ of the input example as the objective function to train the model:
\begin{equation}
    \mathcal{L} = -\log P(l_1,l_2,...,l_N|W)
\end{equation}
where $L = l_1,l_2,...,l_N$ is the golden label sequence for $W$. Finally, the Viterbi decoder is employed to infer the sequence of labels with highest score for the input sentence.

\begin{figure}
    \centering
    \includegraphics[scale=0.5]{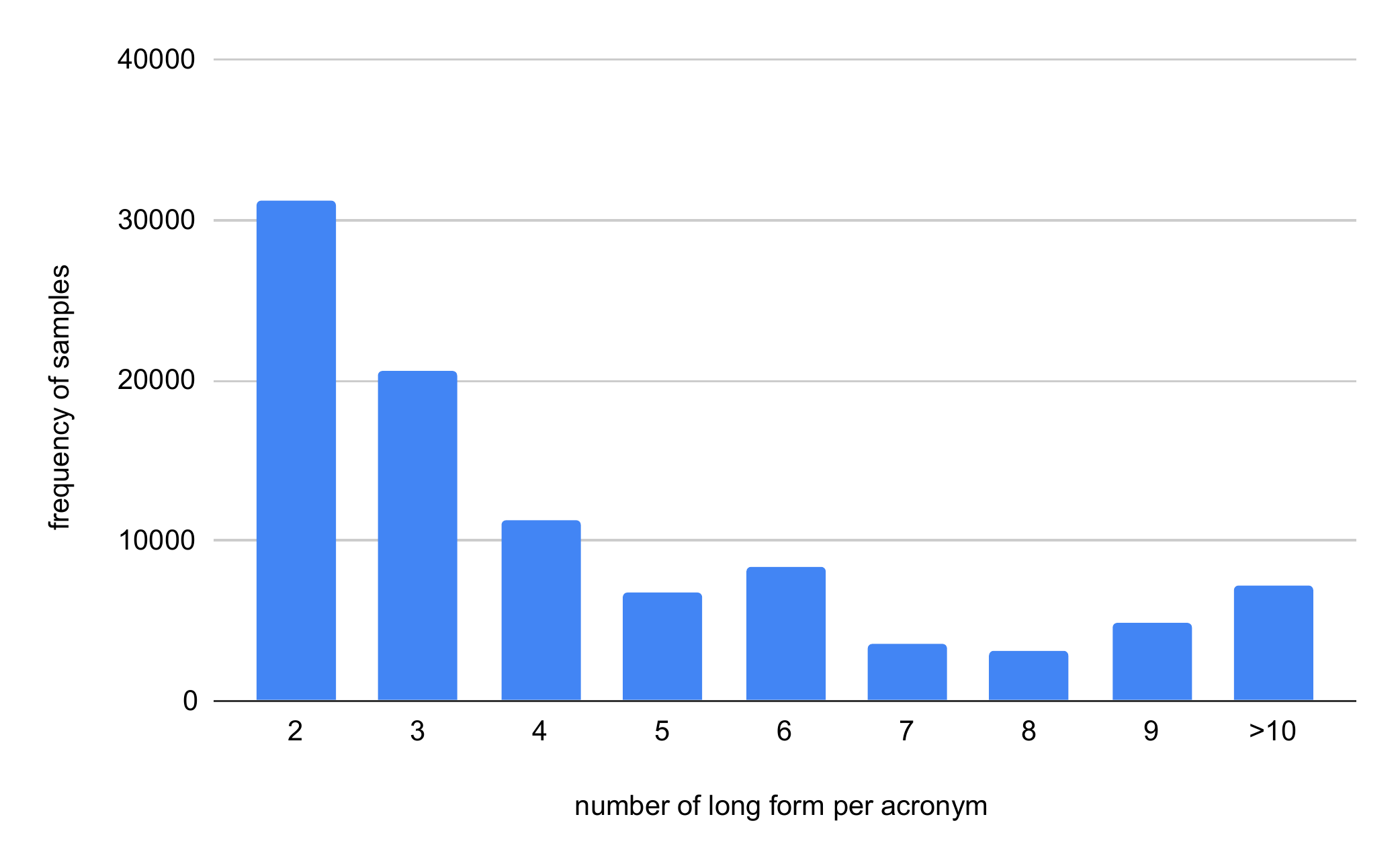}
    \caption{Distribution of samples based on number of long form per acronym}
    \label{fig:freq}
\end{figure}

\section{Hyper Parameters and Dataset Statistics}
\label{sec:parameters}
\begin{table}[]
    \centering
    \begin{tabular}{l|c|c|c}
        Dataset & Train & Dev. & Test \\ \hline
        SciAI & 14,006 & 1,750 & 1,750 \\
        SciAD & 49,788 & 6,233 & 6,225 \\
        UAD & 459,869 & 57,484 & 57,484 \\
        SciUAD & 15,982 & 1,997 & 1,999
    \end{tabular}
    \caption{Dataset Statistics. Numbers of samples per each data splits}
    \label{tab:statistics}
\end{table}
\subsection{Hyper Parameters}
In our experiments, for the rule-based, feature-based and language-model-based models, i.e., NOA, ADE, UAD, BIOADI and LNCRF, we use the same hyper parameters and features reported in the original papers. For the deep learning models, i.e., LSTM-CRF, BEM, DECBAE and GAD, we fine tune the hyper parameters using the performance on the development set for each task. More specifically, we find the following hyper parameters for the deep learning models: 50 samples per each mini-batch; 200 hidden dimensions for all feed forward and BiLSTM layers; dropout rate 0.2; two layers of BiLSTM and GCN; and Adam optimizer with learning rate 0.3. Note that for pre-trained word embeddings we use the uncased version of BERT$_{base}$ \cite{devlin2018bert} with 768 dimensions. 

\begin{figure}
    \centering
    \includegraphics[scale=0.5]{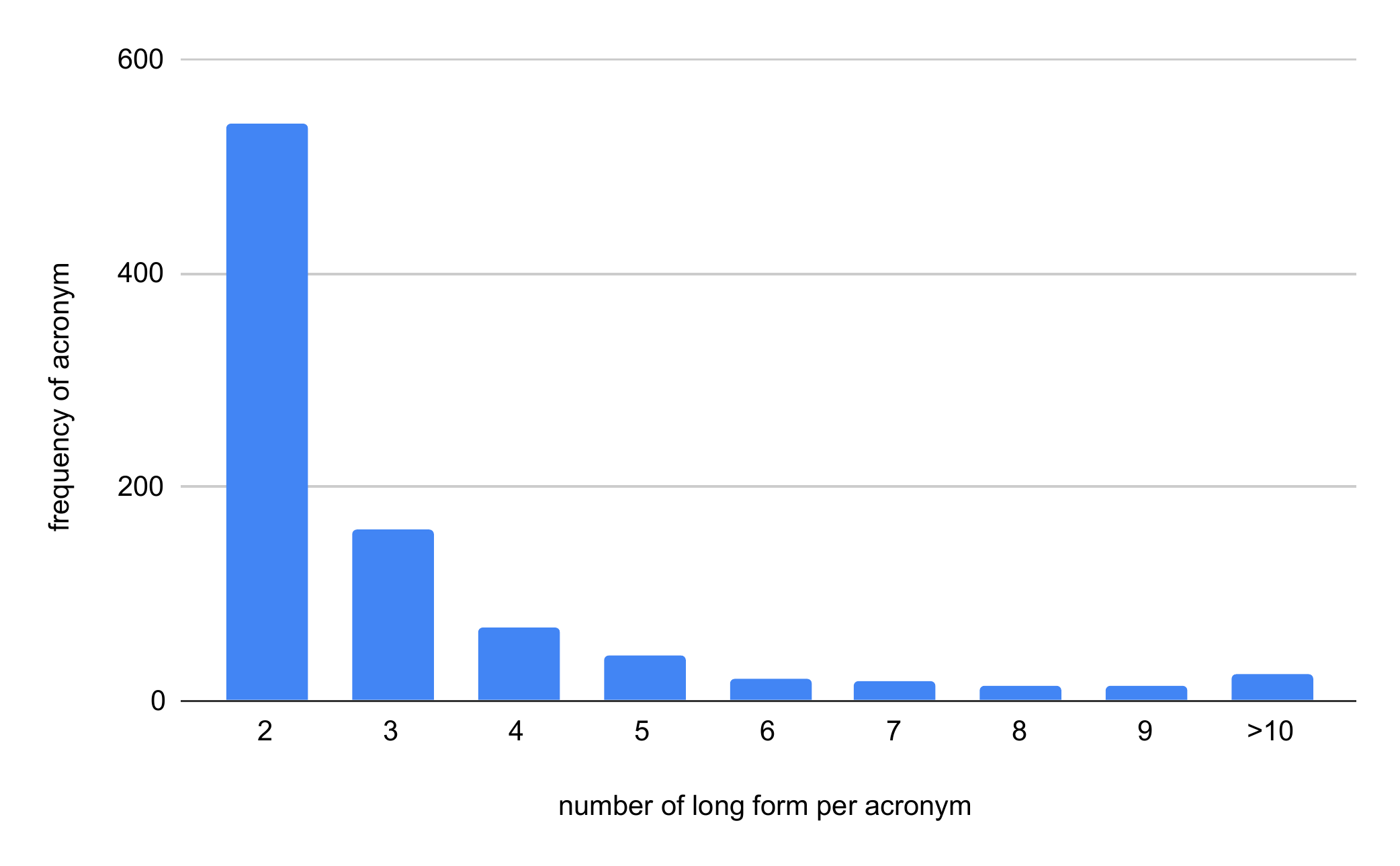}
    \caption{Distribution of acronyms based on number of long form per acronym}
    \label{fig:freq_acr}
\end{figure}

\subsection{Dataset Statistics}
\label{dataset-statistics}
\begin{figure}
    \centering
    \includegraphics[scale=0.5]{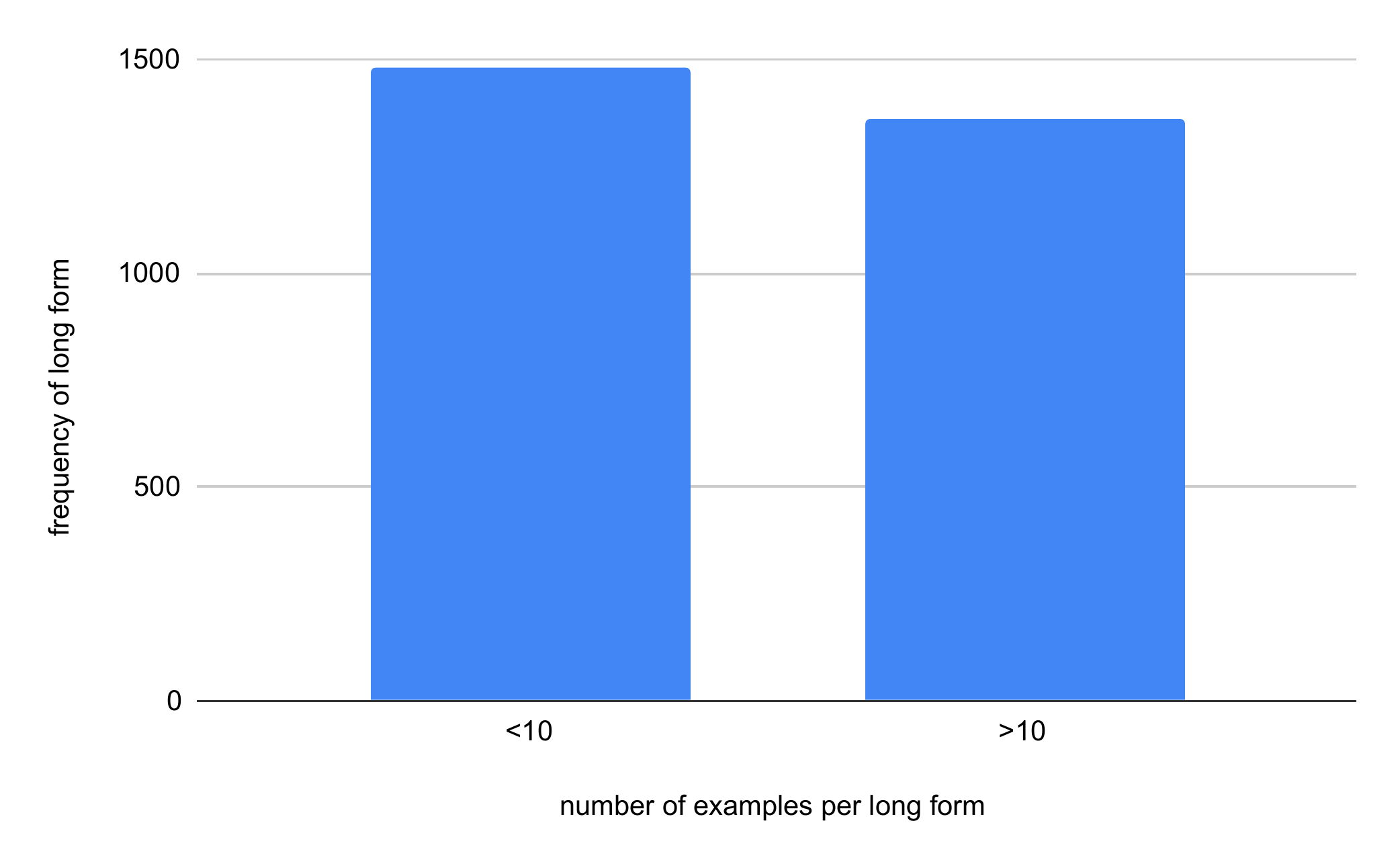}
    \caption{Distribution of long forms based on number of samples per long form}
    \label{fig:freq_meaning}
\end{figure}

For statistics of the datasets used in our experiments, see Table \ref{tab:statistics}. This table shows the number of samples per different data splits of all datasets used in our experiments. In addition, Figures \ref{fig:freq}, \ref{fig:freq_acr} and \ref{fig:freq_meaning} demonstrate more statistics of SciAD dataset. More specifically, Figure \ref{fig:freq} shows the distribution of number of samples based on number of long form per acronym. The distribution shown in this figure is consistent with the same distribution presented in the prior work \cite{charbonnier2018using} in which in both distributions, acronyms with 2 or 3 meanings have the highest number of samples in the dataset. Figure \ref{fig:freq_acr} shows the distribution of number of acronyms based on number of long forms per acronym. Again, this distribution is compatible with the distribution of samples, i.e., Figure \ref{fig:freq}, as acronyms with 2 or 3 meanings have the highest frequency in the dataset. Finally, Figure \ref{fig:freq_meaning}, depicts the number of long forms with either less or more than 10 samples in the dataset. This figure shows that the number of high-frequent and low-frequent long forms are on par with each other in our dataset, so it provides opportunity for future research to train and evaluate acronym disambiguation models on both categories of long forms (i.e., high-frequent and low-frequent long forms)

\subsection{Comparison with the Existing AI and AD Datasets}
\label{dataset-comparison}
This sections compares SciAI and SciAD with the existing acronym identification and disambiguation datasets. Table \ref{tab:compare-AI} compares SciAI with the manually labeled non-medical acronym identification datasets proposed and studied by the prior work. This table shows that our dataset is substantially larger than the existing datasets in various criteria, including the number of sentences present in the dataset and the number of distinct acronyms and long forms. Moreover, we publicly release SciAI to provide more research opportunity for acronym identification in scientific domain. For acronym disambiguation, we compare our model with the existing AD datasets in scientific domain. The results are shown in Table \ref{tab:compare-AD}. This table shows that our proposed dataset is considerably larger than the existing datasets. Especially considering the average number of samples per long form, our dataset is more suitable for advanced deep learning models that is also corroborated in our experiments. 

\begin{figure}
    \centering
    \includegraphics[scale=0.35]{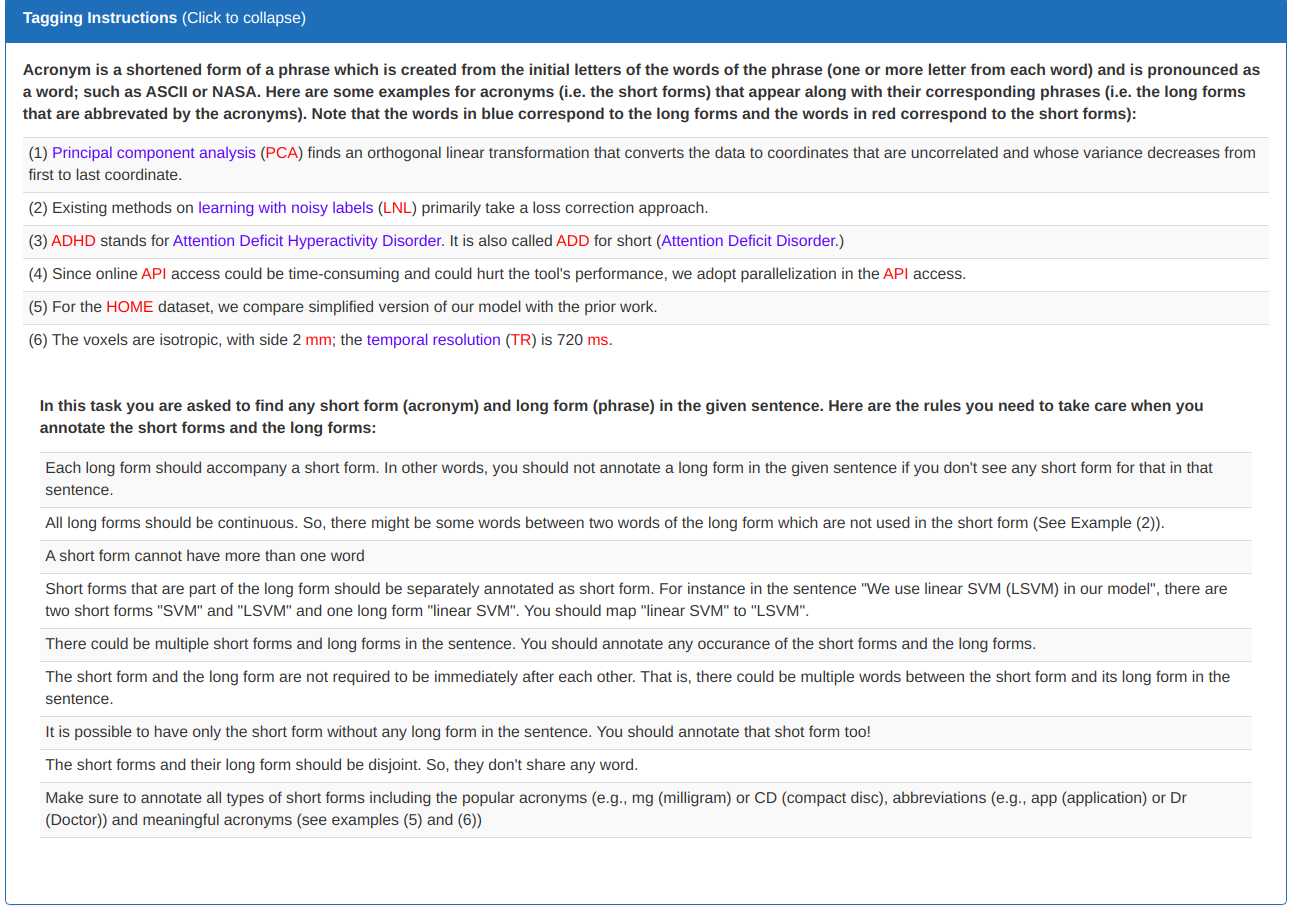}
    \caption{Annotation instructions shown to the annotators} 
    \label{fig:instructions}
\end{figure}

\begin{figure}
    \centering
    \includegraphics[scale=0.3]{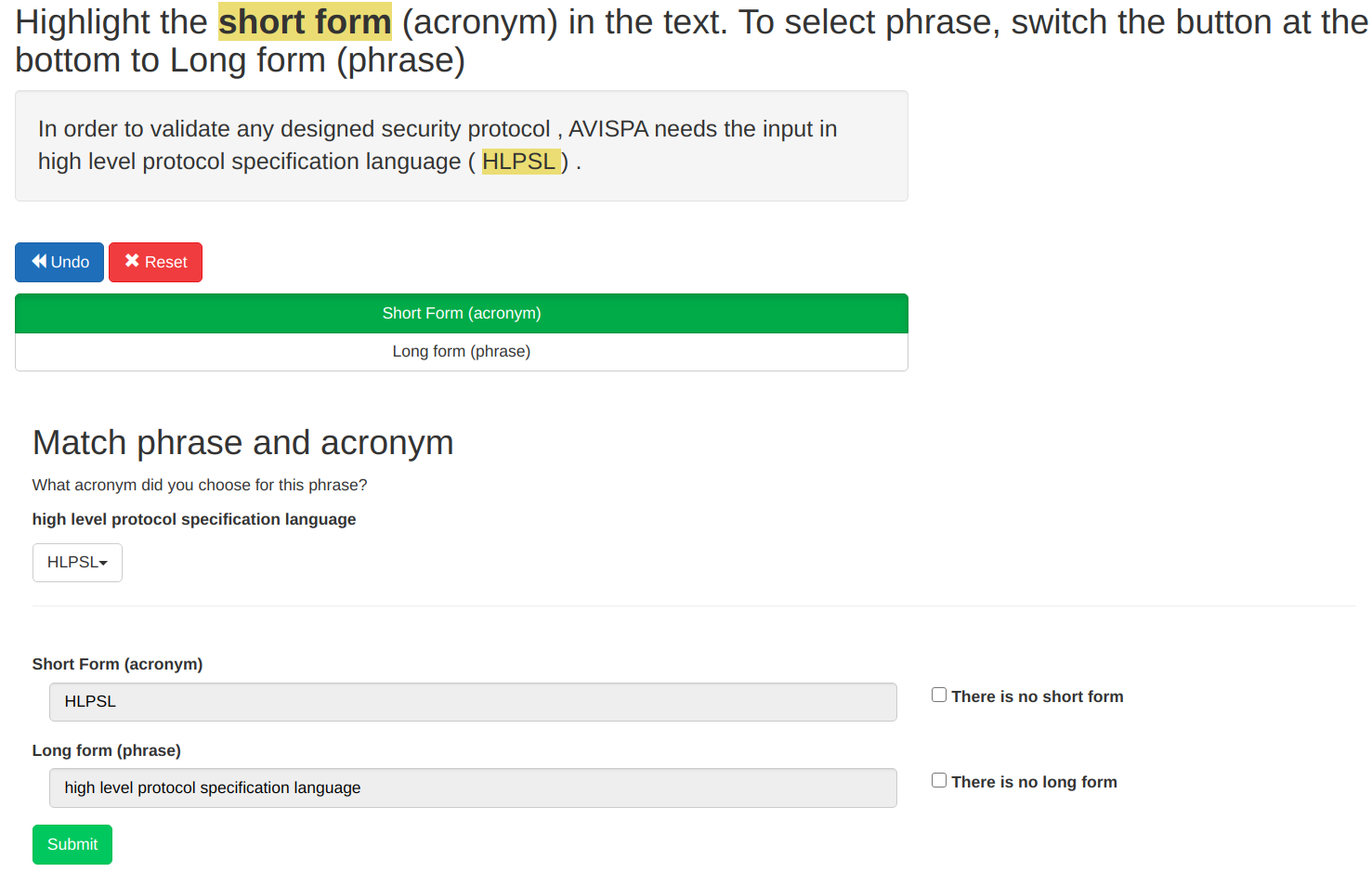}
    \caption{An example of annotation interface. } 
    \label{fig:example}
\end{figure}

\begin{table*}[ht]
    \centering
    \resizebox{.95\textwidth}{!}{
    \begin{tabular}{l|c|c|c|c|c|c}
        Dataset & Size & \# Unique Acronyms & \# Unique Meaning & \# Documents & Publicly Available & Domain \\ \hline
        LSAEF \cite{liu2011learning} & 6,185 & 255 & 1,372 & N/A & No & Wikipedia \\
        AESM \cite{nautial2014finding} & 355 & N/A & N/A & N/A & No & Wikipedia \\
        MHIR \cite{harris2019my} & N/A & N/A & N/A & 50 & No & Scientific Papers \\
        MHIR \cite{harris2019my} & N/A & N/A & N/A & 50 & No & Patent \\
        MHIR \cite{harris2019my} & N/A & N/A & N/A & 50 & No & News \\
        SciAE (ours) & 17,506 & 7,964 & 9,775 & 6,786 & yes & Scientific Papers
    \end{tabular}
    }
    \caption{Comparison of non-medical manually annotated acronym identification datasets. Note that size refers to the number of sentences in the dataset.}
    \label{tab:compare-AI}
\end{table*}

\begin{table*}[ht]
    \centering
    \resizebox{.89\textwidth}{!}{
    \begin{tabular}{l|c|c|c}
        Dataset & Size & Annotation & Avg. Number of Samples per Long Form \\ \hline
        Science WISE \cite{prokofyev2013ontology} & 5,217 & Disambiguation mannualy annotated & N/A \\
        NOA \cite{charbonnier2018using} & 19,954 & No manual annotation & 4 \\
        SciAD (ours) & 62,441 & Acronym identification manually annotated & 22
    \end{tabular}
    }
    \caption{Comparison of scientific acronym disambiguation (AD) datasets. Note that size refers to the number of sentences in the dataset.}
    \label{tab:compare-AD}
\end{table*}

\section{Annotation Instructions and Interface}
\label{sec:interface}
Annotation instruction is shown in Figure \ref{fig:instructions}. This instruction is shown directly above the annotation interface so that the annotators could access it before attempting to annotate. Note that while we ask the annotators to annotate embedded acronyms (e.g., ``\textit{svm}" in ``\textit{linear svm (LSVM)}"), but in the released dataset we keep the long form label for the embedded acronyms to make it consistent with sequence labeling framework for AI. In addition to the instruction shown to the annotators, we also require the annotators to pass a qualification test with the same interface. Only the annotators with 100\% score from the qualification test are allowed to accept the actual HITs.

An example of annotation interface is shown in Figure \ref{fig:example}. In this example the annotator has selected ``\textit{high level protocol specification language}" as the long form and ``\textit{HLPSL}" as the short form (i.e., shown at the bottom of the page). In addition, the annotator has mapped the short form ``\textit{HLPSL}" to the long form ``\textit{high level protocol specification language}" (shown in the middle section). Note that when the annotators switch the button from ``\textit{short forms (acronyms)}" to ``\textit{Long Forms (phrases)}" (i.e., the green buttons at the top of the page), instead of the short forms, the long forms will be highlighted in the text box.


\begin{figure}
    \centering
    \includegraphics[scale=0.25]{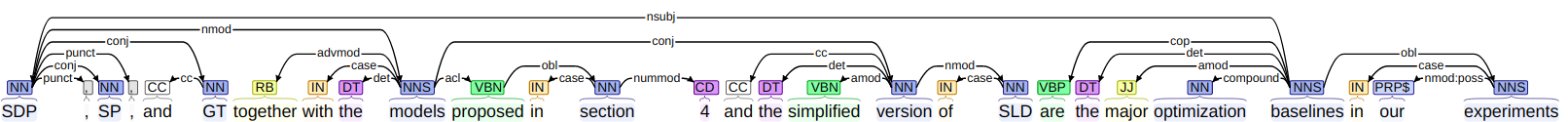}
    \caption{Dependency tree of the sentence ``\textit{SDP, SP, and GT together with the models proposed in section 4  and the simplified version of SLD are the major optimization baselines in our experiments}"}
    \label{fig:dep}
\end{figure}



\end{document}